\documentclass{article}

\usepackage[linesnumbered,boxed,ruled,commentsnumbered]{algorithm2e}
\usepackage{PRIMEarxiv}
\usepackage{amsmath}
\usepackage[utf8]{inputenc} 
\usepackage[T1]{fontenc}    
\usepackage{hyperref}       
\usepackage{url}            
\usepackage{booktabs}       
\usepackage{amsfonts}       
\usepackage{nicefrac}       
\usepackage{microtype}      
\usepackage{lipsum}
\usepackage{fancyhdr}       
\usepackage{graphicx}       
\graphicspath{{media/}}     

\title{Semi-supervised 2D Human Pose Estimation via Adaptive Keypoint Masking
}

\author{
  Kexin Meng \\
  Beijing University of Chemical Technology \\
  Beijing\\
  \texttt{2021210461@buct.edu.cn} \\
   \And
  Ruirui Li \\
  Beijing University of Chemical Technology \\
  Beijing\\
  \texttt{liruirui@mail.buct.edu.cn} \\
   \AND
  Daguang Jiang \\
  Beijing University of Chemical Technology \\
  Beijing\\
  \texttt{jdg@mail.buct.edu.cn} \\
}

\begin{document}
\maketitle

\begin{abstract}
Human pose estimation is a fundamental and challenging task in computer vision. Larger-scale and more accurate keypoint annotations, while helpful for improving the accuracy of supervised pose estimation, are often expensive and difficult to obtain. Semi-supervised pose estimation tries to leverage a large amount of unlabeled data to improve model performance, which can alleviate the problem of insufficient labeled samples. The latest semi-supervised learning usually adopts a strong and weak data augmented teacher-student learning framework to deal with the challenge of "Human postural diversity and its long-tailed distribution". Appropriate data augmentation method is one of the key factors affecting the accuracy and generalization of semi-supervised models. Aiming at the problem that the difference of sample learning is not considered in the fixed keypoint masking augmentation method, this paper proposes an adaptive keypoint masking method, which can fully mine the information in the samples and obtain better estimation performance. In order to further improve the generalization and robustness of the model, this paper proposes a dual-branch data augmentation scheme, which can perform Mixup on samples and features on the basis of adaptive keypoint masking. The effectiveness of the proposed method is verified on COCO and MPII, outperforming the state-of-the-art semi-supervised pose estimation by 5.2$\%$ and 0.3$\%$, respectively.
\end{abstract}


\section{Introduction}
Human pose estimation is a fundamental and challenging task in computer vision. Many practical applications, including behavior recognition, human-computer interaction, and motion capture, are based on human pose estimation tasks. 2D human pose estimation is to identify and estimate the keypoints of the human bodies from a single RGB image. The accuracy of these keypoints is crucial for pose description and behavior analysis. Google proposed DeepPose \cite{b1} in 2014, the first deep neural networks for human body pose estimation. Since then, deep learning-based models have developed rapidly. Although they have significantly improved the accuracy and generalization ability of human pose estimation, there is still room for improvement in accuracy due to the large variability of human poses and the long-tailed distribution. Accurate models rely on large and accurately annotated datasets containing a wide variety of human poses. Unfortunately, such datasets are difficult to obtain. On the one hand, the cost of manual labeling is high, and on the other hand, automatic human body keypoint labeling contains noise. To solve this problem, one idea is to build robust learning against noisy labels, and another idea is to achieve equivalent accuracy by performing semi-supervised learning on a small number of accurate labels.

Semi-supervised learning uses a large amount of unlabeled data to improve the performance of the model. It reduces the reliance on annotations and is more suitable for real-world applications. A new research direction in deep semi-supervised learning is to use unlabeled data to strengthen the training model so that it meets the clustering assumption that the learned decision boundary must lie in a low-density region. Inspired by this idea, the research work of Xie $et$ $al.$ \cite{b2} applied the teacher-student model to 2D pose estimation for the first time. This work investigates the model collapse problem encountered in the field of semi-supervised 2D pose estimation, and designs an effective data augmentation method based on keypoint masking. However, this method has two obvious weaknesses. Firstly, the keypoint masking data augmentation method proposed by Xie $et$ $al.$ does not take into account the difference in sample difficulty. It simply employs a fixed amount of random data augmentation, which may lead to a decrease in model generalization. For keypoint estimation of easily-learned regular poses, it is possible that fewer keypoint annotations are enough. On the contrary, for uncommon poses, more keypoints need to be retained so that it can optimize the recognition of these poses through supervised learning. Furthermore, even for the same pose, the learning difficulty of different keypoints is different. Randomly choosing the amount of masking keypoints may cause some poses or some keypoints to be consistently inaccurately estimated. Second, the model is not robust to attacks, as it does not leverage the smoothness assumption and manifold assumption in semi-supervised learning to design or add a suitable Mixup \cite{b29} augmentation strategy to improve the generalization of the model.

To address above issues, this paper designs an adaptive masking mechanism for keypoints from the perspective of optimization. This method estimates the difficulty of the body pose by analyzing the posterior probability of keypoints, so as to dynamically adjust the number of masks. On the basis of adaptive keypoint masking, this paper also adds the Mixup data augmentation to respond to the smoothness assumption, and adds the Mixup feature enhancement to respond to the Manifold assumption. We verify the method proposed in this paper on the COCO \cite{b3} and MPII\cite{b4} public datasets. On the COCO dataset, where we use 1K labeled data, our method improves the average accuracy by 5.2$\%$. On the MPII dataset, we use MPII training set as labeled data and AI challeger dataset as unlabeled data. Our method improves the average accuracy by 0.3$\%$. In summary, our approach has three main contributions:

\begin{itemize}
\item For 2D pose data augmentation, we propose an adaptive keypoint masking method. It estimates the learning difficulty for different samples and sets different numbers of keypoint masks so that hard samples have more chances to be learned.
\item In order to maximize the utilization of unlabeled data, based on the teacher-student semi-supervised learning framework, we design and use a two-branch strong augmentation method. One is image rotation and adaptive keypoint masking; the other is random Mixup at the sample and feature levels. The combination of the two augmentation methods can further improve the generalization of the model.
\item We fully experiment the proposed method on two public datasets COCO and MPII. Experiments show that the proposed data augmentation can provide more accurate and robust results.
\end{itemize}

\begin{figure}[htbp]
\centerline{\includegraphics[width=0.9\textwidth,height=0.55\textwidth]{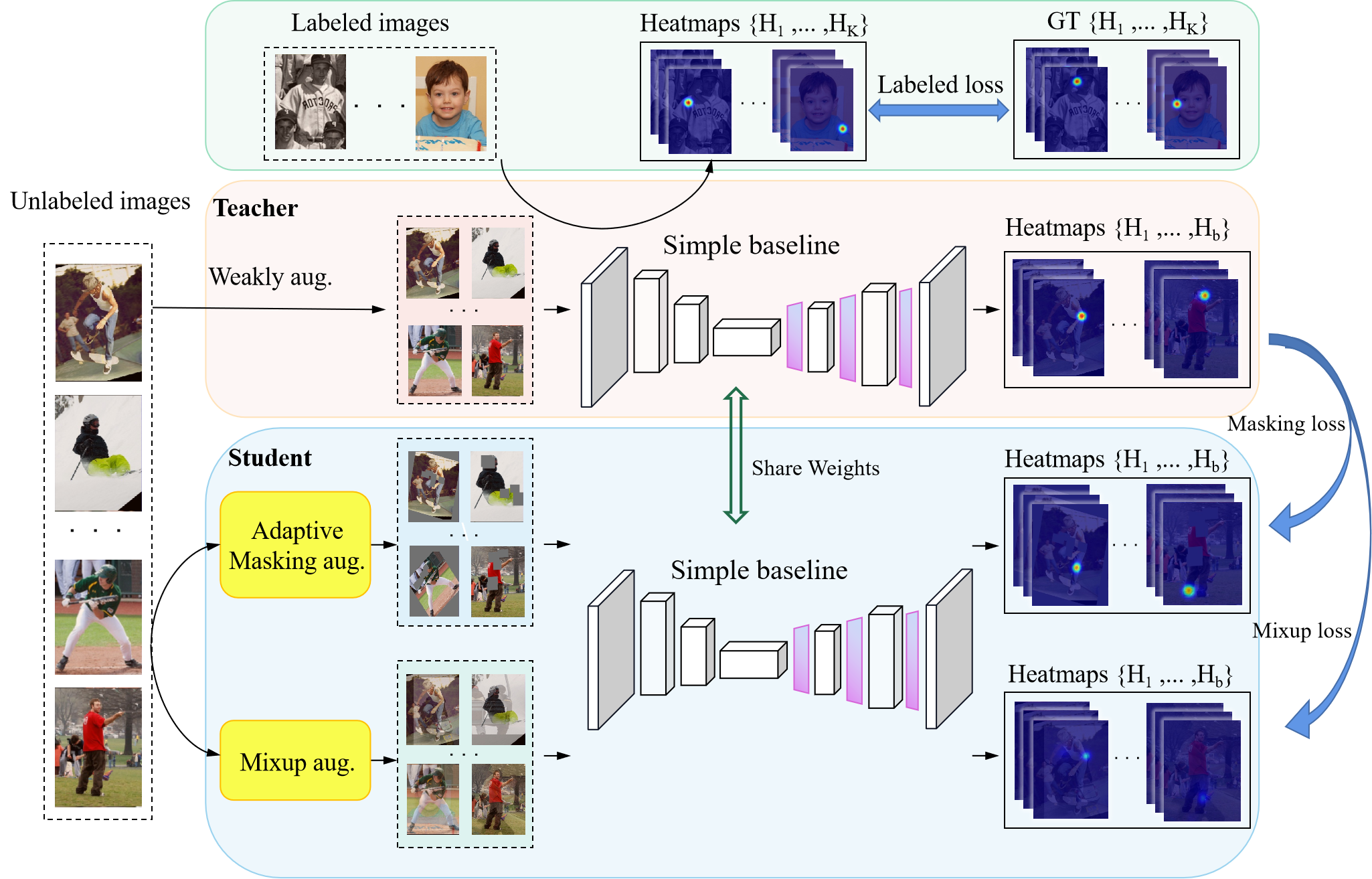}}
\caption{\textbf{Overall overview of our Semi-supervised 2D Human Estimaiton method.} The green part represents the process of supervised training using labeled data, which generates heatmaps through label-guided supervision. The red and blue parts refer to the unsupervised training process based on the teacher-student framework. The red part serves as the teacher, and the heatmaps estimated from weakly augmented unlabeled data guide the learning of the student part. The blue part acts as the student, and we propose two novel data augmentation approaches: Adaptive Keypoint Masking augmentation and Mixup augmentation. These two strong augmentation components are supervised by pseudo heatmaps. The supervised part and the unsupervised teacher-student part share the same network parameters.}
\label{all}
\end{figure}

\section{Related Works}
\label{sec:headings}

\subsection{2D Human Pose Estimation}

The goal of 2D human pose estimation is to identify and locate the human body keypoints in the picture. These keypoints are connected in order and the pose of the human body is obtained. Pose estimation techniques can be broadly categorized into two major approaches: heatmap-based methods \cite{b5, b6, b7, b8, b9, b10} and coordinate regression methods \cite{b11, b12, b13, b14}. Heatmap-based methods represent the detection of keypoints by utilizing two-dimensional Gaussian distributions centered around their respective positions. In contrast to coordinate-based methods, heatmaps offer spatially enriched supervision information, leading to more robust performance. As a result, heatmap-based methods have emerged as the prevailing approach in the field of 2D human pose estimation.

Sun $et$ $al.$ \cite{b7} introduces a heatmap-based high-resolution network known as HRNet. HRNet maintains a high-resolution representation by concatenating multi-resolution convolutional layers in parallel. Moreover, it enhances the high-resolution representation through iterative multi-scale fusion across the parallel convolutions, ultimately generating a high-resolution heatmap. In our experiments on the MPII dataset, we will utilize HRNet as the foundational backbone network to estimate the heatmaps. Xiao $et$ $al.$ \cite{b8} highlights the remarkable progress achieved in deep learning-based human pose estimation, and he also mentioned that the neural network structure in this field is becoming more and more complex. He proposes a simple but effective baseline to reduce the complexity of the algorithm. His method integrates multiple deconvolution layers into the hourglass residual module \cite{b9}. Experiments show that this simple idea does achieve good results. This method has a simple structure but high accuracy, so we will use this method as the backbone network when training on the COCO dataset.

\subsection{Semi-supervised Learning}

Semi-supervised learning makes full use of unlabeled data and provides a solution to improve model performance under limited labeled data conditions. Existing semi-supervised methods have achieved high recognition accuracy in image classification task. There are two categories for semi-supervised image classification task: pseudo-labelling \cite{b15, b16, b17} and consistency regularization \cite{b18, b19, b20, b21, b22, b23}. The pseudo-labeling method may mislabel unlabeled labels, thus seriously affecting the performance of the model, so will not be discussed here. The consistency regularization method means that the model's prediction of unlabeled images should remain the same even after adding noise.

Recognized as a remarkable advancement in the field of semi-supervised learning, MixMatch \cite{b21} exhibits superior performance on small-scale datasets, surpassing other comparable models. It first augmented the labeled data once, and then augmented the unlabeled data several times. The augmented labeled data and unlabeled data are mixed up to obtain a new dataset. Inspired by this work, we also choose to mix unlabeled samples to improve the robustness of the model. In this work, we respond not only to the smoothness assumption at the sample level, but also to the manifold assumption at the feature level.

FixMatch \cite{b22} adopts a teacher-student framework based on strong-weak augmentation. In this framework, the teacher model is trained using labeled data, while the student model learns from the teacher model to improve its performance on unlabeled data. Through the teacher-student framework, the student model can learn more robust feature representations and classification decisions from the teacher model, thereby improving the robustness and generalization ability. This helps alleviate the problem of insufficient annotated data and improves the accuracy of classification. We have designed a new human pose estimation method based on a semi-supervised teacher-student framework. Our method designs two student networks that assist each other to obtain better performance.

\begin{figure}[htbp]
\centerline{\includegraphics[width=0.6\textwidth,height=0.4\textwidth]{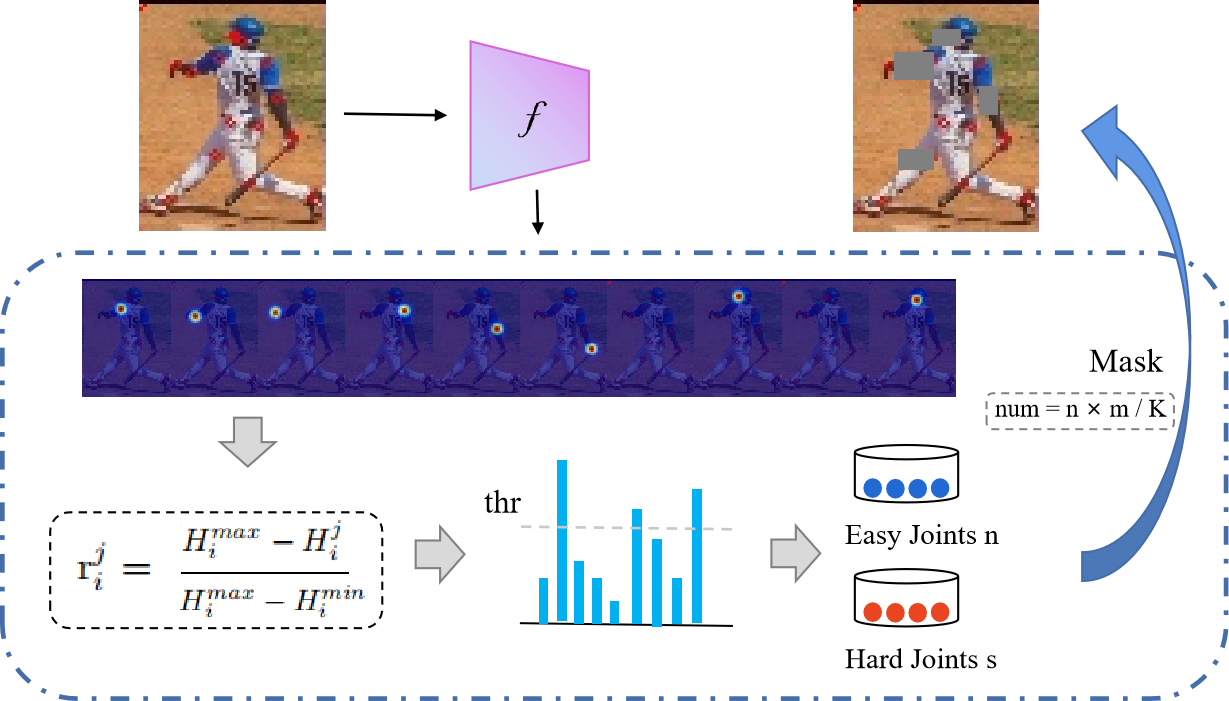}}
\caption{\textbf{The calculation process for allocating the quantity of adaptive keypoint masks.} The weakly augmented image $i$ is passed through the network to produce the heatmaps of the number of keypoints $K$. The maximum value in each heatmap is taken as the heatmap's response $H_{i}^{j}$, and the relative response of each heatmap $r_{i}^{j}$ is calculated based on the formula in the figure. By setting a threshold and using the relative response, the keypoints are divided into two categories: simple and difficult. The quantity of masks allocated to each sample is calculated based on the proportion of keypoints classified as simple among the total number of keypoints. }
\label{adapt}
\end{figure}

Xie $et$ $al.$ \cite{b2} introduced the semi-supervised method into the task of 2D human pose estimation for the first time, and his method is based on consistency regularization. This method uses a teacher-student network where the teacher model is utilized to estimate the unlabeled data and obtain the initial pose estimation result. Subsequently,the accuracy of this initial estimate is enhanced through a series of data augmentation techniques, such as rotation, scaling, and keypoint masking. By training the student model on augmented data, the collapse problem can be alleviated. We further improve the method to make it more suitable for the task of human pose estimation, and the model performance is improved.

\section{\textbf{Method}}

\subsection{\textbf{Preliminary}}\label{AA}
Our approach divides the dataset into a set of labeled $\left\{X_{i}^{l}, H_{i}^{l}\right\}_{i=1}^N$ and unlabeled data $\left\{X_{i}^{u}\right\}_{i=1}^M$. $X$, $N$, and $M$ represent the image, the amount of labeled data, and the amount of unlabeled data, respectively. $H$ in the formula is the ground truth of the heatmap. Many current methods treat pose estimation as a heatmap estimation task, estimating a heatmap for each joint. Within these heatmaps, each pixel value corresponds to the probability of containing body joints. Correspondingly, the position of the maximum value in each heatmap is considered as the position of the keypoint.

Figure \ref{all} illustrates the overall framework of the network. The network is divided into a supervised part and an unsupervised part. In the supervised part, the labeled data is fed into the network, where the objective is to minimize the discrepancy between the estimated heatmaps and the ground truth. In the unsupervised part, the weakly augmented network assumes the role of a teacher, generating pseudo-labels based on its predictions. Conversely, the strongly augmented part acts as a student, aiming to generate predictions that align with the pseudo-labels produced by the teacher. Network parameter sharing is realized between supervised part and unsupervised part. Equation \eqref{eq1} represents our model training loss. $L_{total}$, $L_{s}$, $L_{u}$ and $L_{m}$ are the overall loss, supervised loss and two unsupervised losses of the model, respectively. $\lambda_{u}$ and $\lambda_{m}$ are the weights of the unsupervised losses:

\begin{equation}
L_{total} = L_{s} + \lambda_{u} \cdot L_{u} + \lambda_{m} \cdot L_{m}\label{eq}, 
\end{equation}

\subsection{\textbf{Adaptive Keypoint Masking}}

In the unsupervised keypoint masking augmentation method, the network trying to train first makes a rough estimate of the weakly augmented image. In this way, the approximate positions of the keypoints can be obtained. In the rough estimation of the results, Xie $et$ $al.$ \cite{b2} simply employs a fixed amount of random data augmentation. Taking the selected keypoints as the center, random-sized masks are applied to them. However, we believe that the randomly assigned number of masks is not optimal for the sample. The difficulty of estimating the keypoints of each sample is different. For some images with uncommon actions or blurry images, it is difficult for the network to estimate each keypoint equally. Setting too many keypoint masks in these difficult samples will divert the attention of network learning from the task of pose estimation. For some samples with explicit actions, too few keypoint masks are not enough for learning deep semantics. So we designed an adaptive keypoint masking method, which assigns an appropriate number of masks to each sample according to the difficulty of the sample. Randomly select the keypoints of the distribution number in the graph, and perform a random size mask to avoid overfitting.

Our adaptive keypoint masking method is depicted in Figure \ref{adapt}. During training, each weakly augmented image will get heatmaps of the number of joints after passing through the network. Each pixel value in the heatmap represents the probability of a body keypoint being present at that location. The highest probability value in each heatmap is referred to as its responsiveness, denoted as $H_{i}$. The complete set of responsivities for a single sample is represented as $\left\{H_{1}, ..., H_{K}\right\}$, where $K$ represents the number of keypoints to be estimated. The keypoint heatmap in the sample is normalized for its responsivity based on its distance from the highest responsivity. After finding the heatmap $H_{Max}$ with the largest responsiveness and the heatmap $H_{Min}$ with the smallest responsiveness, the rest of the heatmaps are normalized according to their distance from the highest responsiveness. Thus, the relative responsiveness sequence of all keypoints in a sample is obtained. According to the number of responsivity in the relative response sequence of this sample reaching the specified threshold, the proportion of keypoints that are difficult to estimate in this sample is judged. Thus, this sample is divided into simple sample or difficult sample, and finally an appropriate mask number is assigned to this sample.

There are also some extremely difficult samples in the data, such as very blurry pictures. Their specific performance is that the responsivity of all heatmaps estimated by this sample is very low, and the model cannot recognize the existence of relevant nodes in this picture. For such samples, we pick a minimum threshold for responsiveness. If the highest responsiveness $H_{Max}$ in all heatmaps does not reach this threshold, we classify the sample as an extremely difficult sample. And assign the minimum number of masks set for this sample.

\IncMargin{1em}    
\begin{algorithm}    
\caption{Adaptive keypoint masking algorithm}\label{algorithm}    
\KwData{Weak augmentation $\alpha$, confidence threshold $\gamma$, number of keypoints $K$, unlabeled batch $\left\{x_{1}, \ldots, x_{M}\right\}$, backbone model $f$, Parameter set $m$}    
\For{$i = 1$ \KwTo $M$}{    
    $\triangleright$ Estimation for (weakly augmented) unlabeled data\\  
    $\widetilde{x_{i}} = \alpha(x_{i})$\\  
    $H_{i} = f(\widetilde{x_{i}})$\\  
    $H_{i} = \left\{H_{i}^{1}, \ldots, H_{i}^{K}\right\}$\\  
    $\triangleright$ Calculation of the number of adaptive keypoint masks\\  
    $H_i^{\max} = \max \left\{H_{i}^{1}, \ldots, H_{i}^{K}\right\}$\\  
    $H_i^{\min} = \min \left\{H_{i}^{1}, \ldots, H_{i}^{K}\right\}$\\  
    \For{$j = 1$ \KwTo $K$}{    
        $r_{i}^{j} = \frac{H_{i}^{\max} - H_{i}^{j}}{H_{i}^{\max} - H_{i}^{\min}}$    
    }    
    $n = \sum_{j=1}^{K} \mathbb{I} \left\{r_{i}^{j} < \gamma \right\}$ \\  
    $num = n \times m / K$\\    
}    
\end{algorithm}    
\DecMargin{1em}

This method of adaptive masking can find out keypoint heatmaps whose relative responsivity is greater than a threshold. These heatmaps are on the side closer to maximum responsiveness. The ratio of the number of heatmaps on this side to the number of all keypoints is an indicator of the difficulty of the sample. The larger the ratio, the less difficult the sample is to detect and the more reliable the results are. The smaller the ratio is, the opposite is true.

\begin{figure}[htbp]
\centerline{\includegraphics[width=0.43\textwidth,height=0.32\textwidth]{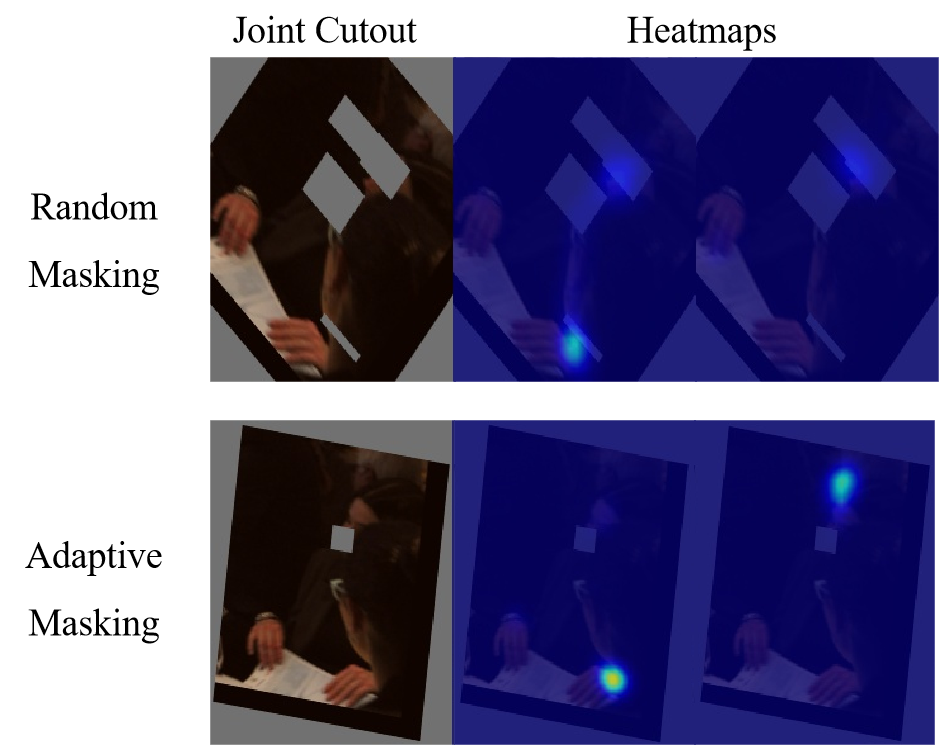}}
\caption{\textbf{The effect of adaptive keypoint masking.} In this example, human poses are indistinguishable. Our method assigns it fewer masks, and the network can acquire more semantic information. Better results can be obtained for keypoint estimation.}
\label{compare}
\end{figure}

\begin{figure}[htbp]
\centerline{\includegraphics[width=0.48\textwidth,height=0.2\textwidth]{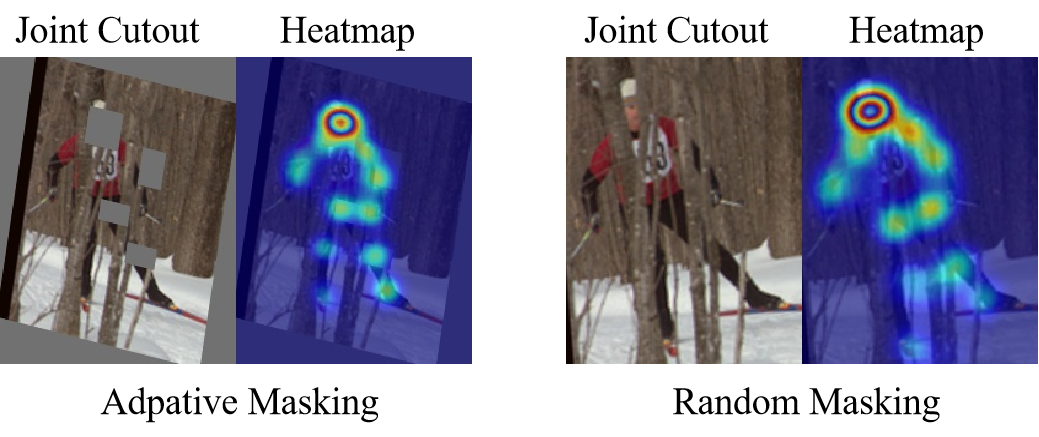}}
\caption{\textbf{The effect of adaptive keypoint masking.} The human pose in this image is more explicit, and our method assigns more keypoint masks to it. It can be seen from the heatmap that the network can mine more deep information based on more masks and obtain better estimation results.}
\label{simple}
\end{figure}

The effect of the method of adaptive keypoint masking is shown in Figure \ref{compare} and Figure \ref{simple}. In Figure \ref{compare}, since the shooting environment is very dark, it is difficult to distinguish the pose of the human body from the picture. For such difficult samples, Xie $et$ $al.$'s method \cite{b2} randomly assigns four masks to it, which makes the model even more unable to obtain useful information, and the pose of the human body has not been learned. Whereas our method divides it into hard samples and assigns the lowest number of masks, the picture is better learned. In Figure \ref{simple}, the pose of the human body is relatively clear, but there are cases where some keypoints are occluded. If we can learn deep semantics, we can infer the positions of the keypoints. The left image represents our approach, which assigns more masks based on the heatmap response of the samples. The network explores deeper information and learns the correlations between keypoints, resulting in a better estimation of the occluded right leg. The image on the right uses Xie $et$ $al.$'s random masking method \cite{b2} with a randomly assigned mask count of 0. The model is in the early stages of learning from the image and cannot accurately infer the positions of the occluded keypoints. These two samples clearly show the effectiveness of our proposed adaptive keypoint masking method.

For simple samples, our adaptive keypoint masking method assigns more masks to them, increasing the intensity of their augmentation, allowing the model to explore deeper semantic information. On the contrary, for difficult samples, the network cannot understand them well during initial training. Our method assigns fewer masks to them, providing the network with more feature information, enabling the network to focus on the basic pose estimation task. After several rounds of training, when difficult samples have been sufficiently learned and the responsiveness of the heatmaps increases, our adaptive masking method will assign them more masks.

\subsection{A Mixing Augmentation Approach to Improve Accuracy}

The semi-supervised network structure consists of one weakly augmented line and two strongly augmented lines. They share network parameters. The heatmaps estimated by the weakly augmented network of unlabeled data are used as pseudo-labels to supervise the results estimated by the two strong augmented lines. The previous part mentioned that an appropriate number of masks for the keypoints can enable the model to learn the internal connection between the keypoints. Even if some of the keypoints are covered, the model can still infer the structural connection between the keypoints to make a relatively correct estimation. However, at the early stage of network training, the performance of the model is unstable and the estimation of the keypoint locations for the samples is not accurate enough. This may lead to poor masking effect on the keypoints of the sample. Therefore, our work also proposes to add a mixed augmentation route, the addition of which can alleviate the bias of the keypoint masking method and greatly improve the accuracy of the estimation.

\textbf{Mixup} The noise samples in the data may affect the performance of the model, and most deep neural network models are trained to minimize the average error of the model on the training set. The trained model may perform poorly in the face of adversarial samples. The proposal of Mixup can alleviate these phenomena. As a simple data augmentation method that is not related to the data set, Mixup can make the model's estimation of the sample less absolute. This can effectively reduce training overfitting. The construction of the virtual sample based on Mixup is shown in Equation \eqref{eq2}.

\begin{gather}
x = \alpha x_{i} + (1 - \alpha) x_{j}, \label{eq2}\\
y = \alpha y_{i} + (1 - \alpha) y_{j}. \label{eq3}
\end{gather}

Where $\alpha$ $\in$ [0, 1],$ x_{i}$, $x_{j}$ are the images to be mixed. $y_{i}$ and $y_{j}$ are the labels of the images. $\alpha$ is a proportion of mixing, which is derived from the beta distribution. This method takes the sample $x_{i}$ of $\alpha$ ratio, and the sample $x_{j}$ of 1-$\alpha$ ratio to fuse into a new sample. As shown in Equation \eqref{eq3}, Mixup not only mixes samples, but also mixes labels accordingly. The labels corresponding to the mixed samples are mixed in the same proportion to obtain a label of the new sample.

This method can expand the number of samples by mixing pictures with other pictures in the same batch at the input layer. Our work not only performs mixing at the image level, but also extends to the feature level of a random layer in the network. We feed unlabeled data into the network and forward propagate to a randomly selected layer to obtain hidden feature representations. We extract the feature representation, and perform pairwise interpolation calculations on the sample features in a batch to generate new sample features. The mixed features are forward-propagated to get the output. Thus far, our method responds to the manifold assumption, increasing the certainty of the model.

The calculation of the loss function is shown in Equation \eqref{eq4}, which is similar to the sample mixing process and also requires linear interpolation. Instead of mixing the pseudo heatmaps to get new labels, we choose to add mixing to the calculation of the loss function. We mix a sample $X_{i}$ with $X_{j}$ or a layer of features $\widetilde{X}_{i}$ with $\widetilde{X}_{j}$ in the same batch to get $X_{m}$ or $\widetilde{X}_{m}$. In the weak augmentation route, we perform pose estimation on $X_{i}$ and $X_{j}$, obtain pseudo heatmaps $H_{i}$ and $H_{j}$. The heatmap estimated by $X_{m}$ or $\widetilde{X}_{m}$ and the pseudo heatmap $H_{i}$ and $H_{j}$ calculate the loss respectively, and then add them according to the ratio of sample mixing:

\begin{equation}
L_{m} = \alpha \times E|| f(X_{m})-H_{i} ||^{2} + (1 - \alpha) \times E|| f(X_{m})-H_{j} ||^{2},\label{eq4}
\end{equation}

\section{Experiment}
\subsection{Baselines and Our Methods}

We introduce semi-supervised classification methods to pose estimation tasks, such as pseudo-label methods. We also list Xie $et$ $al.$'s method \cite{b2} and our method for comparison.

\textbf{Supervised} This is a supervised 2D human pose estimation method. The network structure of this method is very simple. It only inserts several layers of deconvolution in ResNet to expand the low-resolution feature map to the original image size. Finally, the heatmaps needed to predict the keypoints are generated. It exists as a simple baseline for pose estimation methods.

\textbf{PseudoPose} It is changing from pseudo labeling methods. We first train the model with labeled data, then fix the network, and use the estimated results of unlabeled data as pseudo-labels.

\textbf{Single} This is a method based on consistency regularization, which introduces semi-supervised learning into the field of 2D human pose estimation for the first time. It utilizes a teacher-student framework to perform strong and weak augmentations on unlabeled data, aiming to minimize the discrepancy between the estimation results of strong and weak augmentations.

\textbf{Ours} This is a method that can evaluate the difficulty of pose estimation in samples and dynamically adjust the number of keypoint masks according to the difficulty. Compared with ``Single", this method can calculate the appropriate number of masks for each sample based on the heatmap responsivity estimated by the model to be trained.

\textbf{Ours(plus)} After adding the adaptive keypoint masking, the instability of the model may bring bias to the dynamic mask. This may have an impact on the estimation results, so we add a mixing augmentation line.

\subsection{Dataset, Evaluation Metrics and Training}

\begin{table*}[htbp]
\renewcommand{\arraystretch}{1.5}
\caption{Compared with other methods on the COCO dataset. Randomly select a certain amount of data (three sets of experiments) in the training set as labeled data, and the rest as unlabeled data. + means that this augmentation is used in the method.}
\label{1}
\begin{center}
\setlength{\tabcolsep}{2mm}{
\begin{tabular}{|c|c|c|c|c|c|c|c|}
\hline
  & \textbf{Affine} &  & \textbf{Adaptive} &  & \multicolumn{3}{|c|}{\textbf{AP($\%$)}}\\
\cline{6-8} 
 \textbf{Method} & \textbf{Transform} & \textbf{Joint Cutout} & \textbf{Joint Cutout} & \textbf{Mixup} & \textbf{1K labels} & \textbf{5K labels} & \textbf{10K labels} \\
\hline
Supervised & $+$ & & & & 31.5 & 46.4 & 51.1\\
PseudoPose & $+$ & & & & 37.2 & 50.9 & 56.0\\
Single & $+$ & $+$ & & & 42.1 & 52.3 & 57.3\\
\hline
Ours & $+$ & & $+$ & & 42.9 & 53.1 & 57.8\\
\textbf{Ours(plus)} & $+$ & & $+$ & $+$ & \textbf{47.3} & \textbf{56.1} & \textbf{59.1}\\
\hline
\end{tabular}}
\end{center}
\end{table*}

\textbf{COCO} The COCO dataset contains labels of 17 keypoints of 250K human bodies and more than 200K images. We train our model on COCO train2017 with more than 100K images. The training uses Simple Baseline as our network and Resnet18 as the network backbone. We evaluate our method with different number of label settings. We set three kinds of data composition, randomly select 1K, 5K and 10K data in the training set as labeled data for training, and use all the remaining data as unlabeled data. The input image size is 256×192.

\textbf{MPII} It has about 25K images, including 40K human targets marked with information of 16 keypoints. We set the training set of this dataset as labeled data, and use all the data of the AI Challenger dataset \cite{b24} as our unlabeled data. The AI Challenger dataset has about 370K human instances. The size of the input image is set to 256×256.

\textbf{Evaluation metric} The standard evaluation metric for training results on the COCO dataset is based on object keypoint similarity (OKS). $OKS = \Sigma_{i}exp\left\{-d^{2}_{p^{2}}/2S^{2}_{p}\sigma^{2}_{i}\right\}\delta(v_{p^{i}}=1) /
 \Sigma_{i}\delta(v_{p^{i}}=1)$, $d_{p}$ is the Euclidean distance between the detected keypoint and the true value keypoint, $\sigma_{i}$ represents the normalization factor of the i-th keypoint, which reflects the degree of influence of the current keypoint on the whole. We report mean precision and recall scores: AP50 (OKS = 0.50), AP75, AP (the mean of AP scores at 10 OKS positions, 0.50, 0.55, . . . , 0.90, 0.95). All the methods compared above adopt the same augmentation method as Xie $et$ $al.$ \cite{b2} to prove the effectiveness of the method proposed in this work.

PCK is used for evaluation on MPII and AI Challenger datasets. PCK is defined as the proportion of correctly estimated keypoints. Calculate the ratio of the normalized distance between the detected keypoint and its corresponding ground truth value less than the set threshold. PCK@0.5 means that the set threshold is 0.5. The normalized distance is the Euclidean distance between the predicted value of the keypoint and the manually labeled value, and then the human scale factor is normalized. The MPII dataset uses the current person's head diameter as the scale factor (the Euclidean distance between the upper left point and the lower right point of the head rectangle), and the pose estimation index using this scale factor is also called PCKh.

\textbf{Training} Our training on the COCO dataset uses Simple Baseline and Resnet18 \cite{b25} as the backbone, and for the MPII and AI Challenger datasets we use HRNet-W32 as the backbone. We use Adam \cite{b28} optimizer. In the case of COCO dataset and 1K labeled datas, train for 30 epochs. The initial learning rate is 1$e^{-3}$, and the learning rate drops to 1$e^{-4}$ and 1$e^{-5}$ at epochs 20 and 25. In the case of 5K and 10K labeled datas, the model is trained for 100 epochs, and the learning rate is decreased at 70 and 90 epochs. On the MPII dataset, train for 100 epochs. The initial learning rate is 1$e^{-3}$, and the learning rate drops to 1$e^{-4}$ and 1$e^{-5}$ at epochs 50 and 70.

\subsection{Comparison with SOTA Methods}

Table~\ref{1} shows the comparison results of different methods on the COCO dataset. We randomly select 1K, 5K, and 10K labeled data in the training set as our training settings. The methods of data augmentation include rotation, random number of masks for keypoints, adaptive mask for keypoints and feature mixture augmentation.

As shown in the results, when a supervised method is given a small amount of data labels, the model performs poorly. Even when the number of labels is increased to 10K, the results are still unsatisfactory. However, with the introduction of a semi-supervised method, the estimation accuracy significantly improves.

\begin{table*}[htbp]
\renewcommand{\arraystretch}{1.2}
\caption{Comparisons on the MPII dataset. The training set of the MPII dataset is labeled data, and the AI Challenger dataset is unlabeled data. The size of the input image is 256×256.}
\begin{center}
\setlength{\tabcolsep}{5mm}{
\begin{tabular}{ccccccccc}
\toprule
Method  &  Hea  &  Sho  &  Elb  &  Wri &  Hip  &  Kne  &  Ank  &  Total  \\
\midrule
Wallin $et$ $al.$ \cite{b30} & 97.2 & 96.0 & 90.9 & 85.5 & 89.2 & 86.2 & 82.0 & 90.1\\
Newell $et$ $al.$ \cite{b9} & 98.2 & 96.3 & 91.2 & 87.1 & 90.1 & 87.4 & 83.6 & 90.9\\
Xiao $et$ $al.$ \cite{b8} & 98.5 & 96.6 & 91.9 & 87.6 & 91.1 & 88.1 & 84.1 & 91.5\\
Ke $et$ $al.$ \cite{b26} & 98.5 & 96.8 & 92.7 & 88.4 & 90.6 & 89.4 & 86.3 & 92.1\\
Sun $et$ $al.$ \cite{b7} & 98.6 & 96.9 & 92.8 & 89 & 91.5 & 89 & 85.7 & 92.3\\
Zhang $et$ $al.$ \cite{b27} & 98.6 & 97.0 & 92.8 & 88.8 & 91.7 & 89.8 & 86.6 & 92.5\\
Xie $et$ $al.$ \cite{b2} & 98.7 & 97.3 & 93.7 & 90.2 & 92.0 & 90.3 & 86.5 & 93.0\\
\textbf{Ours} & 98.7 & 97.4 & 94.0 & 91.1 & 92.2 & 90.8 & 86.5 & \textbf{93.3}\\
\bottomrule
\end{tabular}}
\label{tab2}
\end{center}
\end{table*}

\begin{figure}[htbp]
\centerline{\includegraphics[width=0.52\textwidth,height=0.21\textwidth]{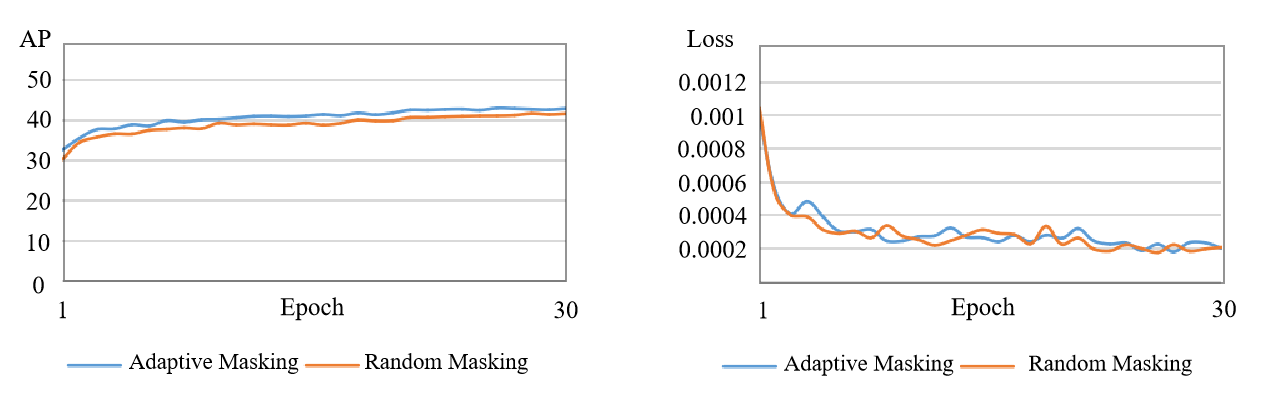}}
\caption{Schematic diagram of AP performance and Training loss of Random Masking method and Adaptive Masking method.}
\label{zhexian}
\end{figure}

In Xie $et$ $al.$'s approach \cite{b2}, the teacher-student semi-supervised framework is introduced into the field of human pose estimation, effectively alleviating the collapsing problem. After incorporating rotation augmentation and random keypoint masking based on coarse heatmaps, the model's performance is further improved, achieving an AP of 42.1$\%$. This result demonstrates that incorporating consistency regularization into the training process can help the model learn meaningful representations in the domain of 2D human pose estimation. Our approach chooses to first evaluate the difficulty of poses in the samples and then assign an appropriate quantity of keypoint masks based on the difficulty. With this more reasonable mask allocation strategy, the accuracy is improved under the three labeled data volume settings. Figure \ref{zhexian} illustrates the AP performance and training loss curves for random keypoint masking method and adaptive keypoint masking method, showing that our method performs better. This shows the effectiveness of adjusting the mask quantity based on sample difficulty.

To prevent deviation in keypoint masking caused by model instability during the early stages of training, we introduced data mix augmentation as a strong augmentation branch. Two strong augmentation branches are supervised by pseudo-labels generated by the teacher. Mixing the data at the image or feature level generates new samples, effectively alleviating model overfitting. Building upon a setting of 1K labeled data, our proposed method achieved a 5.2$\%$ improvement in accuracy compared to Xie $et$ $al.$'s method \cite{b2}, representing the largest improvement among all existing approaches to date.

We also use the training set of the MPII dataset as labeled data and the AI Challenger dataset as unlabeled data for training. Our method uses HRNet-W32 as the backbone. Due to the difference in evaluation metrics used by the COCO dataset, there is also a variation in the performance improvement observed on the two datasets. The experimental results are shown in Table~\ref{tab2}, where our model achieved a performance improvement of 0.3$\%$.

\begin{figure}[htbp]
\centerline{\includegraphics[width=0.47\textwidth,height=0.29   \textwidth]{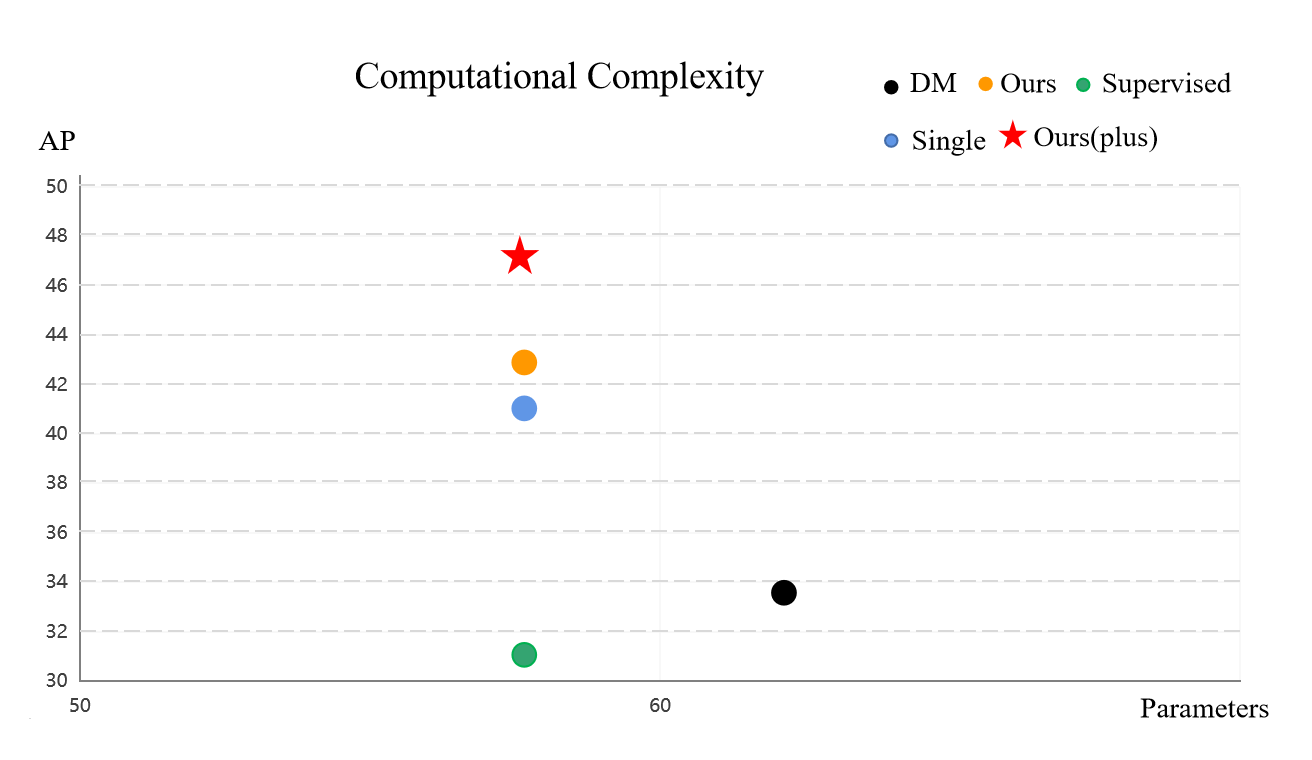}}
\caption{Comparison of the number of parameters among different methods on the COCO dataset.}
\label{Complex}
\end{figure}

When designing a model, the number of parameters should also be taken into consideration. We calculated the number of parameters for different methods on the COCO dataset, and the performance of each method is shown in Figure \ref{Complex}. If only a supervised method is used, there is no reduction in the number of parameters since the backbone is the same as our method, but the performance of the pose estimation is severely degraded. In comparison to Xie $et$ $al.$'s method \cite{b2}, our approach incorporates additional data augmentation modules. However, due to the use of a shared network, the number of parameters remains unchanged while improving accuracy. DM refers to the method proposed by Wallin et al. \cite{b30}, which introduces consistency with an intermediate feature. Despite an increase in parameter count, there is no improvement in accuracy, indicating poor adaptation to pose estimation task.

\subsection{Ablative Study}

\begin{table}[htbp]
\renewcommand{\arraystretch}{1.2}
\caption{The effect of different augmentations of our method on the results. Experiments are performed on three labeled data volume settings on the COCO dataset. "A" represents Affine transformation.}
\begin{center}
\setlength{\tabcolsep}{3.8mm}{
\begin{tabular}{cccc}
\toprule
Augmentation & 1K labels & 5K labels & 10K labels\\
\midrule
A & 38.5 & 50.5 & 55.4\\
A+Adapt & 42.9 & 53.1 & 57.8\\
A+Mixup & 46.0 & 54.8 & 58.7\\
A+Adapt+Mixup & \textbf{47.3} & \textbf{56.1} & \textbf{59.1}\\
\bottomrule
\end{tabular}}
\label{tab3}
\end{center}
\end{table}

We conducted ablation experiments to demonstrate the effectiveness of the proposed data augmentation methods. The experimental results, as shown in Table~\ref{tab3}, indicate that when the network has only one strong augmentation branch, which is rotation and scaling applied to the images, the AP is 38.5$\%$ with 1K labeled data. By incorporating our adaptive keypoint masking into this strong augmentation branch, the performance improved by 4.5$\%$. These results demonstrate that our adaptive keypoint masking method assists the model in achieving more accurate pose estimation by employing a more reasonable masking approach. Furthermore, when the mixed augmentation branch is introduced alone, the accuracy improved by 7.5$\%$. When both of our data augmentation methods are combined, the AP reaches 47.3$\%$ with 1K labeled data. These experimental results indicate that our designed data augmentation methods enhance the model's performance, and when used in conjunction, they significantly assist the model in pose estimation.

\begin{table}[htbp]
\renewcommand{\arraystretch}{1.2}
\caption{The influence of the parameter setting of the dynamic keypoint mask on the experimental results. The experiment is set on the COCO dataset, the number of labels is 1K, and the backbone is Resnet18.}
\begin{center}
\setlength{\tabcolsep}{4mm}{
\begin{tabular}{ccc}
\toprule
MASK JOINT SET & AP($\%$)   & AR($\%$)   \\
\midrule
6 & 41.88   & 45.89   \\
8 & \textbf{42.92} & \textbf{46.74}\\
10 & 42.01 & 46.15\\
12 & 41.70 & 45.61\\
\bottomrule
\end{tabular}}
\label{tab4}
\end{center}
\end{table}

\begin{table}[htbp]
\renewcommand{\arraystretch}{1.2}
\caption{The effect of feature mixing before different layers of the backbone on experimental results.}
\begin{center}
\setlength{\tabcolsep}{4mm}{
\begin{tabular}{ccc}
\toprule
Augmentation & Mixing Location & AP($\%$)\\
\midrule
Affine+Cut+Mix & Before Backbone & 44.67\\
Affine+Cut+Mix & Before Layer-1 & 45.26\\
Affine+Cut+Mix & Before Layer-3 & \textbf{46.22}\\
Affine+Cut+Mix & Before Deconv & 35.47\\
Affine+Cut+Mix & Before Final-layer & 26.57\\
\bottomrule
\end{tabular}}
\label{tab5}
\end{center}
\end{table}

We also conducted a set of comparative experiments to select the most suitable parameters for the adaptive keypoint masking method. The comparative experiments were performed on the COCO dataset with a labeled data quantity of 1K. We chose to add the adaptive keypoint masking method as the sole data augmentation method to the semi-supervised network. The parameters of the method were sequentially set to 6, 8, 10 and 12. The experimental results, as shown in Table~\ref{tab4}, indicate that the model performs best when the parameter is set to 8. With a parameter value of 8, the ratio of the number of keypoint heatmaps with relative responsiveness above the threshold to the total number of keypoints remains within a reasonable range. For samples determined to be easily estimable, a maximum of 8 masks is allocated. For samples deemed extremely difficult, a minimum of 2 masks is assigned. In this way, appropriate data augmentation can be obtained regardless of the difficulty level of the sample.

We propose a hybrid augmentation method and designed experiments to investigate the effectiveness of mixing at different layers of the network. The experimental results are presented in Table~\ref{tab5}. The first experiment involves image-level mixing at the input layer, while the remaining four experiments perform feature-level mixing at different layers of the Simple Baseline network. In the feature-level mixing experiments, image-level mixing is also applied simultaneously. From the experimental results, it is evident that simultaneous random mixing of data at both image and feature levels outperforms mixing at the image level alone. Furthermore, the best results are obtained when feature-level mixing augmentation is applied before $Layer-3$.

\section{Conclusion}

In this work, we propose an adaptive keypoint masking method that utilizes a trainable network to perform coarse analysis on samples, determining the difficulty level of each sample, and dynamically adjusting the number of masks based on their difficulty. This augmentation method assists the network in achieving more accurate semi-supervised learning for human pose estimation. We also propose a better framework for strong-weak augmentation, which combines samples at either the image or feature level to generate a larger variety of new samples. We evaluate our method on two publicly available datasets, and the experimental results demonstrate that our model represents the state-of-the-art performance.

\bibliographystyle{unsrt}  
\bibliography{references}

\begin{thebibliography}{10}

\bibitem{b1}
Alexander Toshev and Christian Szegedy.
\newblock Deeppose: Human pose estimation via deep neural networks.
\newblock In {\em Proceedings of the IEEE conference on computer vision and pattern recognition}, pages 1653--1660, 2014.

\bibitem{b2}
Rongchang Xie, Chunyu Wang, Wenjun Zeng, and Yizhou Wang.
\newblock An empirical study of the collapsing problem in semi-supervised 2d human pose estimation.
\newblock In {\em Proceedings of the IEEE/CVF International Conference on Computer Vision}, pages 11240--11249, 2021.

\bibitem{b29}
Hongyi Zhang, Moustapha Cisse, Yann~N Dauphin, and David Lopez-Paz.
\newblock mixup: Beyond empirical risk minimization.
\newblock {\em arXiv preprint arXiv:1710.09412}, 2017.

\bibitem{b3}
Tsung-Yi Lin, Michael Maire, Serge Belongie, James Hays, Pietro Perona, Deva Ramanan, Piotr Doll{\'a}r, and C~Lawrence Zitnick.
\newblock Microsoft coco: Common objects in context.
\newblock In {\em Computer Vision--ECCV 2014: 13th European Conference, Zurich, Switzerland, September 6-12, 2014, Proceedings, Part V 13}, pages 740--755. Springer, 2014.

\bibitem{b4}
Mykhaylo Andriluka, Leonid Pishchulin, Peter Gehler, and Bernt Schiele.
\newblock 2d human pose estimation: New benchmark and state of the art analysis.
\newblock In {\em Proceedings of the IEEE Conference on computer Vision and Pattern Recognition}, pages 3686--3693, 2014.

\bibitem{b5}
Jonathan~J Tompson, Arjun Jain, Yann LeCun, and Christoph Bregler.
\newblock Joint training of a convolutional network and a graphical model for human pose estimation.
\newblock {\em Advances in neural information processing systems}, 27, 2014.

\bibitem{b6}
Shih-En Wei, Varun Ramakrishna, Takeo Kanade, and Yaser Sheikh.
\newblock Convolutional pose machines.
\newblock In {\em Proceedings of the IEEE conference on Computer Vision and Pattern Recognition}, pages 4724--4732, 2016.

\bibitem{b7}
Ke~Sun, Bin Xiao, Dong Liu, and Jingdong Wang.
\newblock Deep high-resolution representation learning for human pose estimation.
\newblock In {\em Proceedings of the IEEE/CVF conference on computer vision and pattern recognition}, pages 5693--5703, 2019.

\bibitem{b8}
Bin Xiao, Haiping Wu, and Yichen Wei.
\newblock Simple baselines for human pose estimation and tracking.
\newblock In {\em Proceedings of the European conference on computer vision (ECCV)}, pages 466--481, 2018.

\bibitem{b9}
Alejandro Newell, Kaiyu Yang, and Jia Deng.
\newblock Stacked hourglass networks for human pose estimation.
\newblock In {\em Computer Vision--ECCV 2016: 14th European Conference, Amsterdam, The Netherlands, October 11-14, 2016, Proceedings, Part VIII 14}, pages 483--499. Springer, 2016.

\bibitem{b10}
Wei Yang, Wanli Ouyang, Hongsheng Li, and Xiaogang Wang.
\newblock End-to-end learning of deformable mixture of parts and deep convolutional neural networks for human pose estimation.
\newblock In {\em Proceedings of the IEEE conference on computer vision and pattern recognition}, pages 3073--3082, 2016.

\bibitem{b11}
Joao Carreira, Pulkit Agrawal, Katerina Fragkiadaki, and Jitendra Malik.
\newblock Human pose estimation with iterative error feedback.
\newblock In {\em Proceedings of the IEEE conference on computer vision and pattern recognition}, pages 4733--4742, 2016.

\bibitem{b12}
Aiden Nibali, Zhen He, Stuart Morgan, and Luke Prendergast.
\newblock Numerical coordinate regression with convolutional neural networks.
\newblock {\em arXiv preprint arXiv:1801.07372}, 2018.

\bibitem{b13}
Diogo~C Luvizon, Hedi Tabia, and David Picard.
\newblock Human pose regression by combining indirect part detection and contextual information.
\newblock {\em Computers \& Graphics}, 85:15--22, 2019.

\bibitem{b14}
Xiao Sun, Jiaxiang Shang, Shuang Liang, and Yichen Wei.
\newblock Compositional human pose regression.
\newblock In {\em Proceedings of the IEEE international conference on computer vision}, pages 2602--2611, 2017.

\bibitem{b15}
Dong-Hyun Lee et~al.
\newblock Pseudo-label: The simple and efficient semi-supervised learning method for deep neural networks.
\newblock In {\em Workshop on challenges in representation learning, ICML}, volume~3, page 896. Atlanta, 2013.

\bibitem{b16}
Ilija Radosavovic, Piotr Doll{\'a}r, Ross Girshick, Georgia Gkioxari, and Kaiming He.
\newblock Data distillation: Towards omni-supervised learning.
\newblock In {\em Proceedings of the IEEE conference on computer vision and pattern recognition}, pages 4119--4128, 2018.

\bibitem{b17}
Qizhe Xie, Minh-Thang Luong, Eduard Hovy, and Quoc~V Le.
\newblock Self-training with noisy student improves imagenet classification.
\newblock In {\em Proceedings of the IEEE/CVF conference on computer vision and pattern recognition}, pages 10687--10698, 2020.

\bibitem{b18}
Samuli Laine and Timo Aila.
\newblock Temporal ensembling for semi-supervised learning.
\newblock {\em arXiv preprint arXiv:1610.02242}, 2016.

\bibitem{b19}
Mehdi Sajjadi, Mehran Javanmardi, and Tolga Tasdizen.
\newblock Regularization with stochastic transformations and perturbations for deep semi-supervised learning.
\newblock {\em Advances in neural information processing systems}, 29, 2016.

\bibitem{b20}
Antti Tarvainen and Harri Valpola.
\newblock Mean teachers are better role models: Weight-averaged consistency targets improve semi-supervised deep learning results.
\newblock {\em Advances in neural information processing systems}, 30, 2017.

\bibitem{b21}
David Berthelot, Nicholas Carlini, Ian Goodfellow, Nicolas Papernot, Avital Oliver, and Colin~A Raffel.
\newblock Mixmatch: A holistic approach to semi-supervised learning.
\newblock {\em Advances in neural information processing systems}, 32, 2019.

\bibitem{b22}
Kihyuk Sohn, David Berthelot, Nicholas Carlini, Zizhao Zhang, Han Zhang, Colin~A Raffel, Ekin~Dogus Cubuk, Alexey Kurakin, and Chun-Liang Li.
\newblock Fixmatch: Simplifying semi-supervised learning with consistency and confidence.
\newblock {\em Advances in neural information processing systems}, 33:596--608, 2020.

\bibitem{b23}
Jiwon Kim, Youngjo Min, Daehwan Kim, Gyuseong Lee, Junyoung Seo, Kwangrok Ryoo, and Seungryong Kim.
\newblock Conmatch: Semi-supervised learning with confidence-guided consistency regularization.
\newblock In {\em European Conference on Computer Vision}, pages 674--690. Springer, 2022.

\bibitem{b24}
Jiahong Wu, He~Zheng, Bo~Zhao, Yixin Li, Baoming Yan, Rui Liang, Wenjia Wang, Shipei Zhou, Guosen Lin, Yanwei Fu, et~al.
\newblock Large-scale datasets for going deeper in image understanding.
\newblock In {\em 2019 IEEE International Conference on Multimedia and Expo (ICME)}, pages 1480--1485. IEEE, 2019.

\bibitem{b25}
Kaiming He, Xiangyu Zhang, Shaoqing Ren, and Jian Sun.
\newblock Deep residual learning for image recognition.
\newblock In {\em Proceedings of the IEEE conference on computer vision and pattern recognition}, pages 770--778, 2016.

\bibitem{b28}
Diederik~P Kingma and Jimmy Ba.
\newblock Adam: A method for stochastic optimization.
\newblock {\em arXiv preprint arXiv:1412.6980}, 2014.

\bibitem{b30}
Erik Wallin, Lennart Svensson, Fredrik Kahl, and Lars Hammarstrand.
\newblock Doublematch: Improving semi-supervised learning with self-supervision.
\newblock In {\em 2022 26th International Conference on Pattern Recognition (ICPR)}, pages 2871--2877. IEEE, 2022.

\bibitem{b26}
Lipeng Ke, Ming-Ching Chang, Honggang Qi, and Siwei Lyu.
\newblock Multi-scale structure-aware network for human pose estimation.
\newblock In {\em Proceedings of the european conference on computer vision (ECCV)}, pages 713--728, 2018.

\bibitem{b27}
Hong Zhang, Hao Ouyang, Shu Liu, Xiaojuan Qi, Xiaoyong Shen, Ruigang Yang, and Jiaya Jia.
\newblock Human pose estimation with spatial contextual information.
\newblock {\em arXiv preprint arXiv:1901.01760}, 2019.

\end{thebibliography}

\end{document}